\documentclass[10pt,twocolumn,letterpaper]{article}
\pdfoutput=1

\usepackage{iccv}
\usepackage{times}
\usepackage{epsfig}
\usepackage{graphicx}
\usepackage{amsmath}
\usepackage{amssymb}
\usepackage{mathtools}
\usepackage{multicol}
\usepackage{multirow}
\usepackage{appendix}
\usepackage{titling}

    \pretitle{\vspace*{-3pt}\begin{center}\Large \bf}
    \posttitle{\vspace*{14pt}\par\end{center}}
    \preauthor{\large\begin{center}\par}
    \postauthor{\vspace*{4pt}\par\end{center}}
    \predate{}
    \date{}
    \postdate{}

\DeclareMathOperator*{\argmin}{arg\,min}
\DeclareMathOperator{\sign}{sign}

\usepackage{animate} 

\usepackage[super]{nth}
\usepackage{graphicx,subcaption}
\usepackage{floatrow}
\usepackage{xcolor}
\definecolor{darklavender}{rgb}{0.45, 0.31, 0.59}
\definecolor{asparagus}{rgb}{0.53, 0.66, 0.42}
\definecolor{cerulean}{rgb}{0.0, 0.48, 0.65}
\definecolor{outrageousorange}{rgb}{1.0, 0.43, 0.29}
\newcommand*\samethanks[1][\value{footnote}]{\footnotemark[#1]}
\usepackage{hyperref}
\hypersetup{pagebackref=true,breaklinks=true,letterpaper=true,colorlinks,bookmarks=false}

\iccvfinalcopy 

\def\iccvPaperID{5644} 
\def\httilde{\mbox{\tt\raisebox{-.5ex}{\symbol{126}}}}

\ificcvfinal\fi

\begin{document}

\title{3DIAS: 3D Shape Reconstruction with Implicit Algebraic Surfaces}

\author{Mohsen Yavartanoo\thanks{equal contribution} \qquad Jaeyoung Chung\samethanks \qquad Reyhaneh Neshatavar \qquad Kyoung Mu Lee \\ASRI, Department of ECE, Seoul National University, Seoul, Korea\\
{\tt\small \{myavartanoo,robot0321,reyhanehneshat,kyoungmu\}@snu.ac.kr}}

\maketitle
\ificcvfinal\thispagestyle{empty}\fi


\begin{abstract}
3D Shape representation has substantial effects on 3D shape reconstruction. Primitive-based representations approximate a 3D shape mainly by a set of simple implicit primitives, but the low geometrical complexity of the primitives limits the shape resolution. Moreover, setting a sufficient number of primitives for an arbitrary shape is challenging. To overcome these issues, we propose a constrained implicit algebraic surface as the primitive with few learnable coefficients and higher geometrical complexities and a deep neural network to produce these primitives. Our experiments demonstrate the superiorities of our method in terms of representation power compared to the state-of-the-art methods in single RGB image 3D shape reconstruction. Furthermore, we show that our method can semantically learn segments of 3D shapes in an unsupervised manner. The code is publicly available from this  \href{https://myavartanoo.github.io/3dias/}{link}.
\end{abstract}

\section{Introduction}


Single image 3D reconstruction is a procedure of capturing the structure and the surface of 3D shapes from single RGB images, which has various applications in computer vision, computer graphics, computer animation, and augmented reality.
Recent advanced methods have substantially improved 3D shape reconstruction with the advent of deep neural networks (DNNs).
These methods can be mainly categorized based on the representation of 3D shapes into explicit-based ~\cite{brock2016generative, DBLP:journals/corr/AchlioptasDMG17, DBLP:journals/corr/MontiBMRSB16} and implicit-based ~\cite{ONet, deng2020cvxnet, park2019deepsdf, Genova2019LearningST, Superquadrics} methods.
Voxel-grid, as the most straightforward explicit representation, is useful in many applications. However, voxel-based methods generally suffer from large memory usage and quantization artifacts~\cite{Tatarchenko_2017_ICCV}.
Polygon mesh~\cite{DBLP:journals/corr/abs-1803-07549,DBLP:journals/corr/abs-1804-01654} has been introduced as alternative representation. 
However, since many polygon mesh-based methods start from a template mesh and deform it to reconstruct the target 3D shapes~\cite{DBLP:journals/corr/abs-1804-01654, Han2019Imagebased3O}, they can not produce 3D shapes with arbitrary topologies. 

\begin{figure}[!t]
    \includegraphics[width=\linewidth]{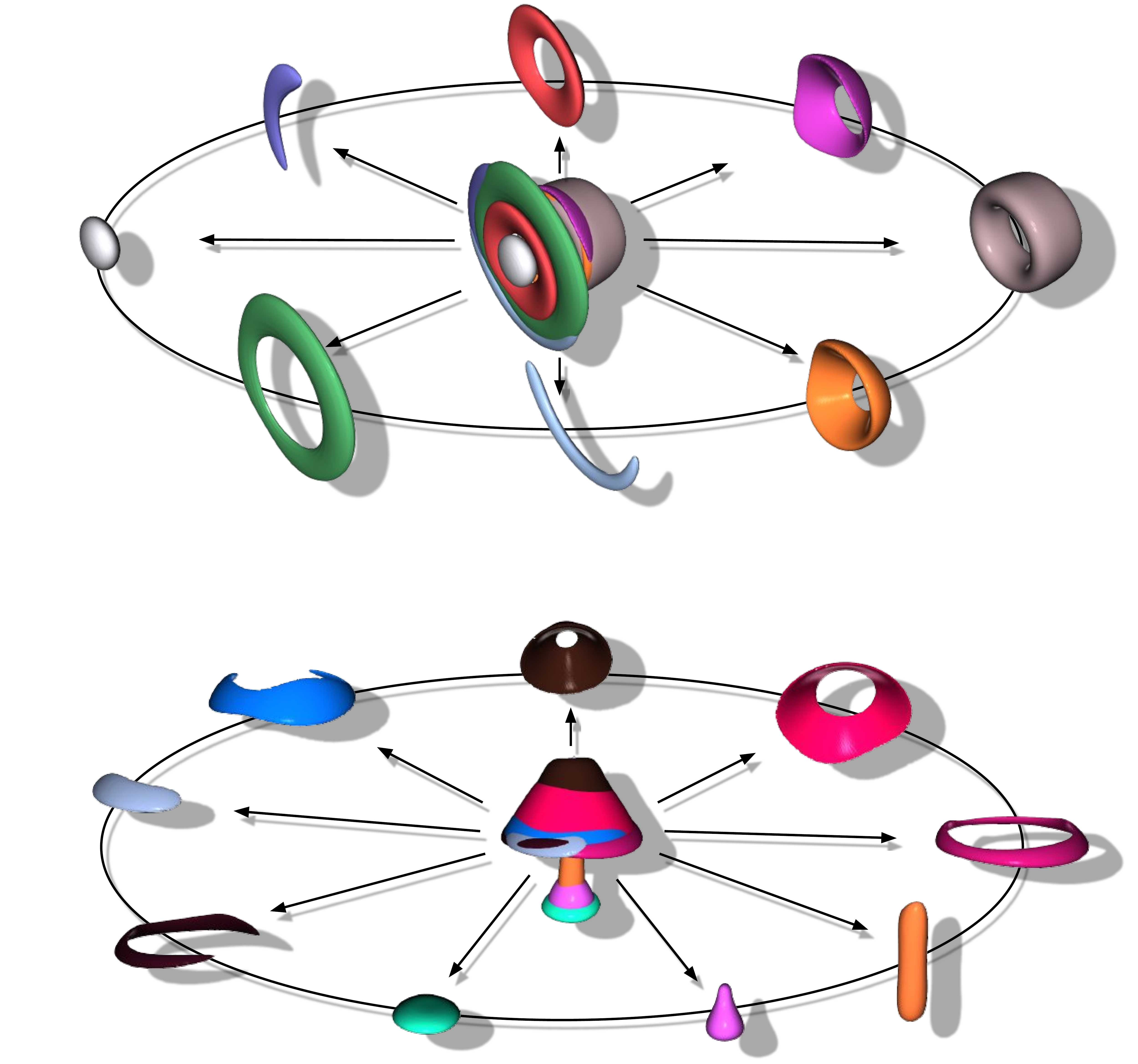}
        \caption{The exploded view of 3DIAS representation. The 3D shapes consist of a union over the proposed constrained implicit algebraic primitives with proper attributes.}
        \label{fig:teaser}
\end{figure}

\begin{figure*}[!t]
        \begin{subfigure}[b]{.209\linewidth}                \includegraphics[width=\linewidth]{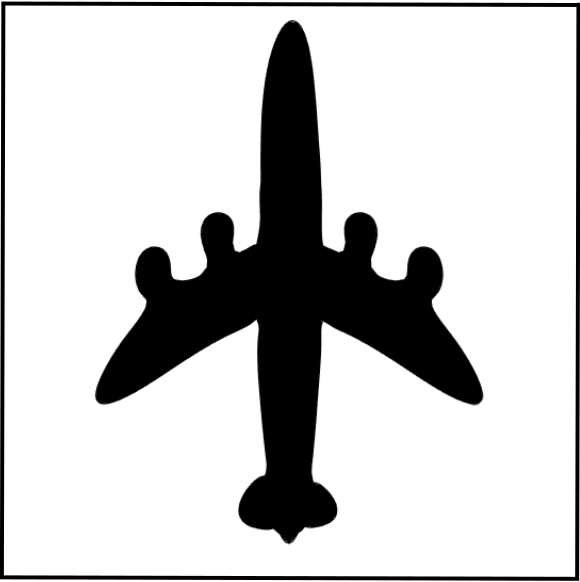}
                \caption{Target}
        \end{subfigure}%
        \qquad
        \begin{subfigure}[b]{.209\linewidth}                \includegraphics[width=\linewidth]{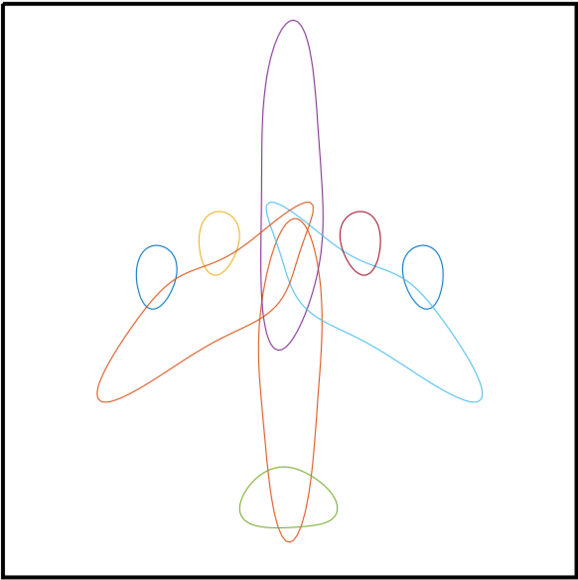}
                \caption{Quartic polynomials}
        \end{subfigure}
        \qquad
        \begin{subfigure}[b]{.209\linewidth}                \includegraphics[width=\linewidth]{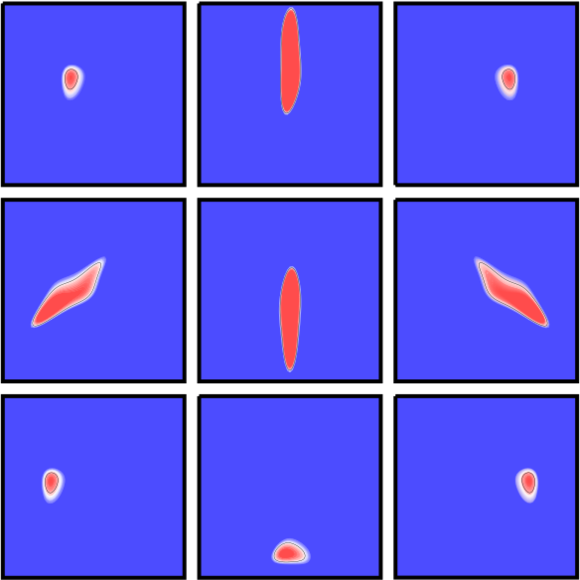}
                \caption{Collection of primitives}
        \end{subfigure}
        \qquad
        \begin{subfigure}[b]{.242\linewidth}                \includegraphics[width=\linewidth]{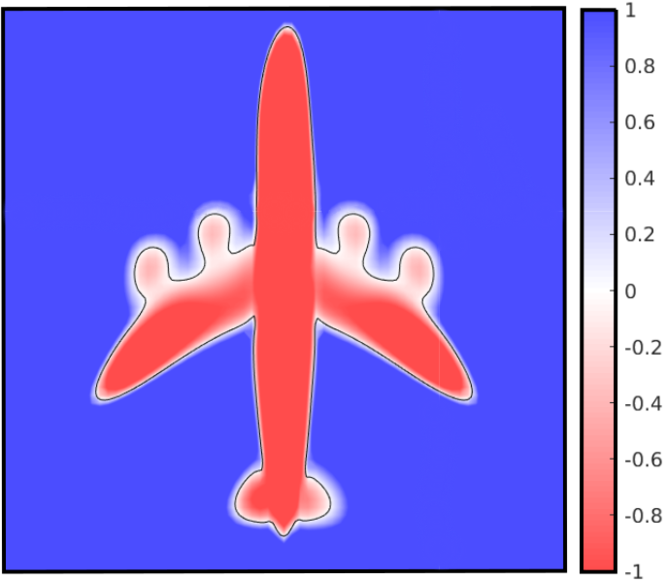}
                \caption{After union}
        \end{subfigure}
        \caption{Composition of implicit algebraic surfaces. Our network approximates the target shape as a union over a set of constrained implicit algebraic surfaces. The network estimates the coefficients and the center of polynomials. Note that since the level sets for each primitive are quite different, the final surface has non-uniform level sets as shown in (d).}\label{Fig:union}
\end{figure*}

On the other hand, implicit representations can approximate surfaces of 3D shapes as zero-sets of continuous functions in the Euclidean space.
Recent implicit-based methods have shown some promises to reconstruct arbitrary shapes without any template. \cite{park2019deepsdf,ONet, Cubes, Genova2019LearningST, Superquadrics, deng2020cvxnet}
These methods can be categorized into two mainstreams; isosurface-based and primitive-based methods. 
Isosurface-based methods generally generate a surface by employing a neural network~\cite{park2019deepsdf,ONet} as an implicit function that assigns negative and positive values or different probabilities to the points lying inside and outside the shape. 
However, for each time visualization, these methods require all the neural network parameters to extract the zero-sets by determining the sign of many sample points in 3D space.
Furthermore, these representations are unsuitable for computer graphics and virtual reality applications because they require additional postprocessing like marching cubes to generate the final 3D shapes. 
Contrastingly, primitive-based methods approximate 3D shapes by a group of primitives such as cubes~\cite{Cubes}, ellipsoids~\cite{Genova2019LearningST}, superquadrics~\cite{Superquadrics}, and convexes~\cite{deng2020cvxnet}.  
Despite their advantages in visualization and direct usage for various applications, the resolutions of reconstructed shapes are limited due to the simple topology (i.e., genus-zero) of the primitives. 
Consequently, approximating a 3D shape requires many of these simple primitives. Moreover, since the geometrical complexity varies from shape to shape, determining a sufficient number of primitives is challenging for an arbitrary shape. 

In this paper, we propose a novel primitive-based 3D shape representation based on the learnable implicit algebraic surfaces named 3DIAS as shown in Figure~\ref{fig:teaser}.
Since implicit algebraic surfaces have high degrees of freedom, they can describe complex shapes better than simple primitives~\cite{Bajaj1992TheEO}.
Besides, identifying an implicit algebraic primitive is straightforward and depends on only a few parameters.
We apply various constraints on these primitives to facilitate learning and achieve detailed appearances. 
We limit our primitives to the class of algebraically solvable implicit algebraic surfaces to assist fast 2D rendering and 3D visualization, which can be useful in many computer graphics applications.
Furthermore, we develop an upper bound constraint with an efficient parameterization to guarantee that the primitives have closed surfaces and controlled sizes.
Finally, we guide the primitives to cover different segments of a target shape by restricting the locations of their centers.  
To generate these primitives, we design a DNN-based encoder-decoder that captures the information of an observation (e.g., single image) and provides the parameters of the primitives.
In our experiments, we show that our method outperforms state-of-the-art methods with most of the metrics. 
Moreover, we experimentally demonstrate that 3DIAS can semantically learn the components of 3D shapes without any supervision and adjust the number of primitives by excluding the primitives with empty volumes.

We summarize, our main \textbf{contributions} as follows:
\begin{itemize}
    \item We propose a novel primitive-based 3D shape representation with the learnable implicit algebraic surfaces, which can produce more complex topologies with few parameters hence appropriate for describing geometrically complex shapes.

    \item We develop various constraints to produce solvable and closed primitives with proper scales in desired locations to ease learning and generate appealing results. 
    
    \item We experimentally demonstrate that 3DIAS outperforms state-of-the-art methods. Furthermore, we show that it can semantically learn the components of 3D shapes and adjust the number of used primitives.  
\end{itemize}

\begin{figure*}[!t]
\begin{center}
\includegraphics[width=17.5cm]{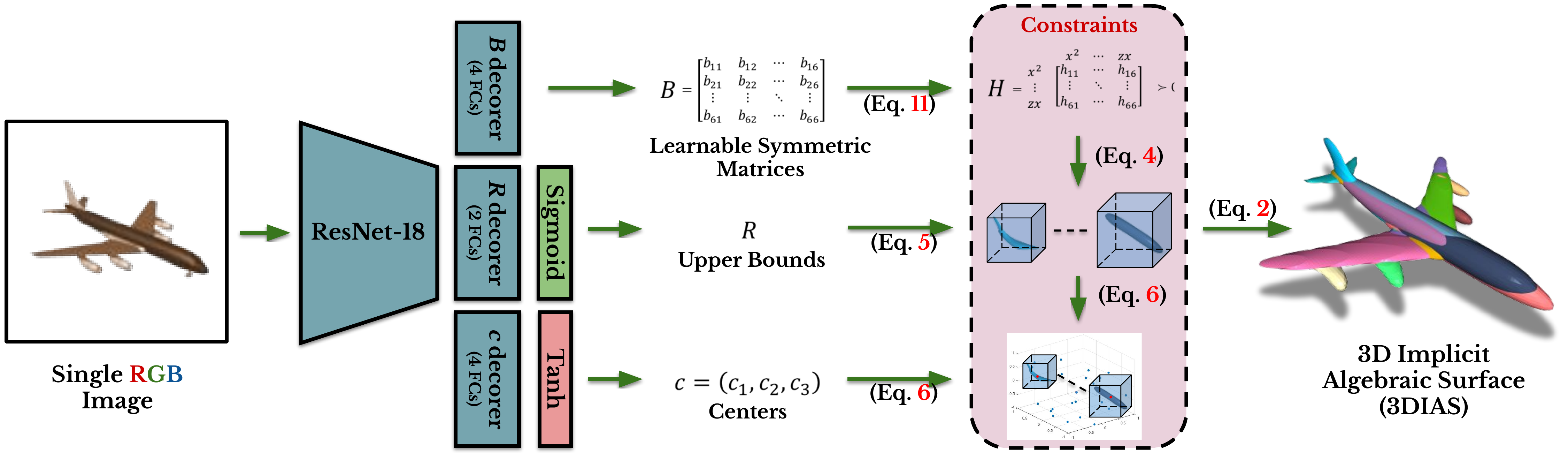}
\end{center}
   \caption{The overview of Single RGB 3D surface reconstruction using 3DIAS. We use an encoder (ResNet-18~\cite{ResNet}) to learn the local and global information from the given single RGB image. Then three sets of fully connected layers decode the latent features to provide the coefficients, the scales, and the location of centers for $M$ primitives. Note that we apply $\min$ operator to take the union over the $M$ primitives.
   }
   \label{Fig:net}
\end{figure*}

\section{Related Work}
In this section, we review some related DNN-based 3D shape reconstruction methods with various representations. 

\noindent
\textbf{Explicit representations.} 
A set of voxel is commonly used for discriminative~\cite{7353481,DBLP:journals/corr/QiSNDYG16} and generative~\cite{DBLP:journals/corr/ChoyXGCS16,DBLP:journals/corr/GirdharFRG16} tasks since it is the most simple way in 3D representation. However, the results represented with voxel have a limitation in resolution due to memory issues. Although ~\cite{DBLP:journals/corr/HaneTM17,inproceedings} proposed to reconstruct 3D objects in a multi-scale fashion, they are still limited to comparably small $256^{3}$ voxel grids and require multiple forward passes to generate final 3D voxels. 
3D point clouds give an alternative representation of 3D shapes. Qi \etal~\cite{DBLP:journals/corr/QiSMG16,DBLP:journals/corr/QiYSG17} pioneered point clouds as a discriminative representation for 3D tasks using deep learning. Fan \etal~\cite{DBLP:journals/corr/FanSG16} introduced point clouds in 3D reconstruction task as an output representation. However, since point clouds have no information for connections between the points, it needs additional post-processing procedures~\cite{article,article2} to build the 3D mesh as an output.
Mesh is another commonly used representation for 3D shape~\cite{groueix2018,DBLP:journals/corr/abs-1712-06584}. However, most of methods in 3D shape reconstruction using meshes generate meshes with simple topology~\cite{DBLP:journals/corr/abs-1804-01654} or utilize a template as a reference. They can only manage the objects from the same class~\cite{DBLP:journals/corr/abs-1712-06584,8100077}
and can not guarantee to produce closed surfaces~\cite{groueix2018}.

\noindent
\textbf{Implicit Representations.} Implicit representation is a good alternative to avoid the problems above. In contrast to the mesh-based approaches, implicit-based methods do not require a template from the same object class as input. There are mainly two approaches; 
isosurface-based and primitive-based methods. 
Chen \etal~\cite{chen2019learning} propose a neural network that takes the 3D points and latent code of the shape, then outputs the values for each point, indicating whether the point is outside or inside the shape as an occupancy function. Park \etal~\cite{park2019deepsdf} utilize the signed distance function to obtain the zero-set surface of the shape.
Tatarchenko \etal~\cite{Tatarchenko_2019_CVPR} proposed the occupancy function that implicitly describes the 3D surfaces as the continuous decision boundary of a DNN-based classifier. Compared to voxel representation, this method can estimate an occupancy function continuous in 3D with a learnable neural network that can generate at any arbitrary resolution.  This approach significantly decreases memory usage during training.  The surface can be extracted as a mesh representation from the learned model at test time by a multi-resolution isosurface extraction method. The finite resolution of the voxel or octree cells limits the accuracy of the reconstructed shape by these methods and their ability to capture fine details of 3D shapes.
Deng \etal~\cite{deng2020cvxnet} represented a shape with a convex combination of half-planes. These methods use a single global latent vector to represent entire surfaces of a 3D shape. The latent vector is decoded into continuous surfaces with the corresponding implicit networks. While this technique successfully models geometry, it often requires many primitives to obtain a desirable appearance, and it is unclear how many primitives are required.

For the 3D reconstruction task, we compare our implicit-based approach against several state-of-the-art implicit-based methods such as Structured Implicit Function (SIF) ~\cite{Genova2019LearningST}, OccNet ~\cite{ONet}, and CvxNet~\cite{deng2020cvxnet}. Moreover, we select Pixel2Mesh~\cite{DBLP:journals/corr/abs-1804-01654}  and AtlasNet~\cite{DBLP:journals/corr/abs-1802-05384} , which use explicit surface generation in contrast to the previous methods.

\section{3DIAS}
In this section, we first introduce our 3D shape representation based on implicit algebraic surfaces. Then we explain the additive constraints for effective learning. Next, we describe the proposed network and learning procedure to reconstruct the surface of 3D shapes with our representation. 

\subsection{3D shape representation}\label{section:3D_shape_reprsentation}
\subsubsection{Implicit algebraic primitives}
We build a complex target 3D shape with a combination of primitives $p(x,y,z)$ that are the building blocks of 3D (i.e., basic geometric forms) as shown in Figure~\ref{fig:teaser}. 
To select a primitive with a large degree of freedom (i.e., complex geometry and topology) and few parameters, we employ the implicit algebraic surface that is a zero-level set of a multivariate polynomial function of $x$, $y$, and $z$ as
Eq.~\ref{Eq:algebraic}:
\begin{equation}\label{Eq:algebraic}
\begin{split}
    p(x,y,z)&=\sum_{\mathclap{0\leq i+j+k\leq d}} a_{ijk}x^iy^jz^k\\
    &= \textit{\textbf{v}}A\textit{\textbf{v}}^T=0,
\end{split}
\end{equation}
where $\textit{\textbf{v}}=[1,x,y,z,x^2,y^2,z^2,\dots]$, and $d$, $a_{ijk}$, and $A$ are the degree, the coefficients, and coefficient matrix of the polynomial function, respectively.
Like many other implicit surfaces, the implicit algebraic surface divides the space and maps points in 3D space into negative and positive values.
Therefore, to represent detailed surfaces $\mathcal{S}(x,y,z)$ of 3D shapes, we can combine these primitives by utilizing constructive solid geometry~\cite{DBLP:books/daglib/0036997} and apply boolean operations to them, which can be formulated as Eq.~\ref{Eq:union}:
\begin{equation}\label{Eq:union}
\begin{split}
    \mathcal{S}(x,y,z)&=\bigcup_{m=1}^{M} p_m(x,y,z)\\
            &=\min (p_1(x,y,z),\dots,p_M(x,y,z)),
\end{split}
\end{equation}
where $M$ is the number of primitives in the union.

\subsubsection{Constraints on primitives}
We apply a set of constraints on the defined implicit algebraic primitive $p(x,y,z)$ to better approximate the surface $\mathcal{S}(x,y,z)$ of a target 3D shape.

\noindent
\textbf{Solvability of primitives.}
Easy visualization and rendering attributes for a 3D shape representation can be useful in many computer graphics and virtual reality applications. The class of implicit algebraic primitives with algebraic solutions are appropriate representations for ray-tracing hence achieving these properties. 
Accordingly, we use multivariate quartic ($d=4$) polynomial functions as the primitives $p(x,y,z)$ because they have the highest degree of freedom among all implicit algebraic primitives with closed-form algebraic solutions~\cite{Galois}.

\noindent
\textbf{Closedness and scales of primitives.}
Beyond the aforementioned constraint, we need to guarantee that the reconstructed shape and hence all primitives have closed surfaces as Figure~\ref{Fig:union}.
We can ensure that a quartic primitive has a closed surface by enforcing its fourth-degree terms to always be positive~\cite{273718} as Eq.~\ref{Eq:closedness}:
\begin{equation}\label{Eq:closedness}
\begin{split}
    p^4(x,y,z) &= \sum_{i+j+k=4} a_{ijk}x^iy^jz^k \\
    & = \textit{\textbf{u}}A_{[5:10]}\textit{\textbf{u}}^T>0,
\end{split}
\end{equation}
where $\textit{\textbf{u}}=[x^2,y^2,z^2,xy,yz,zx]$ and $A_{[5:10]}$ is the $6\times6$ sub-matrix of $A$, including the coefficients of fourth-degree terms.
This implies that $A_{[5:10]}\succ 0$ is a positive definite (PD) matrix.
Note that, with the PD matrix $A_{[5:10]}\succ 0$, the algebraic surface exists if and only if $p^{3\downarrow}(x,y,z)=\sum_{0\leq i+j+k\leq 3}a_{ijk}x^iy^jz^k$ is negative and $|p^{3\downarrow}(x,y,z)|>|p^{4}(x,y,z)|$ for some points in $\mathbb{R}^3$. Otherwise, the primitive has zero volume because it has no real-valued solution.

Moreover, since each primitive reconstructs a different segment of a target 3D shape, we need to ensure that its volume is smaller than the target shape. 
To prevent generating large primitives and control their scales, we develop an upper bound for each primitive. 
To reconstruct a primitive $p(x,y,z)$ included in the upper bound $q(x,y,z)$, it is sufficient to satisfies the inequality $q(x,y,z)<p(x,y,z)$ in $\mathbb{R}^3$ which is also equivalent to Eq.\ref{Eq:inequality}:
\begin{equation}\label{Eq:inequality}
\begin{split}
    p(x,y,z) = h(x,y,z)+q(x,y,z),
\end{split}
\end{equation}
where $h(x,y,z)$ is a positive-valued function. 
The function $h(x,y,z)$ is always positive if and only if its matrix of coefficients $H$ be positive definite. 
As a result, the coefficient matrix $A$ of primitive $p(x,y,z)$ is the summation of a positive definite matrix $H_{10\times10}$ and the coefficient matrix $Q_{10\times10}$ of the upper bound $q(x,y,z)$.
For simplicity we consider the upper bound $q(x,y,z)$ as Eq.\ref{Eq:bound_primitive}:
\begin{equation}\label{Eq:bound_primitive}
\begin{split}
    q(x,y,z) = x^4+y^4+z^4-R,
\end{split}
\end{equation}
where $R$ is a positive value that controls the size of the upper bound.
Note that the developed upper bound is a more general constraint that also holds the criteria for the closedness constraint. For proof, refer to the supplementary materials.

\noindent
\textbf{Locations of primitives.} 
We also encourage the primitives to better cover different components of 3D shape by applying a constraint on their locations. Accordingly, we restrict the locations of their centers $c=(c_1,c_2,c_3)$ into the areas nearby the shape and reformulate the primitives as
Eq.~\ref{Eq:center}:
\begin{equation}\label{Eq:center}
    p(x,y,z)=\sum_{\mathclap{0\leq i+j+k\leq 4}} a_{ijk}(x-c_1)^i(y-c_2)^j(z-c_3)^k=0.
\end{equation}

Therefore, within these constraints, we can reconstruct primitives with controlled scales and locations, which facilitates the reconstruction and provides more details.


\subsection{3D shape reconstruction}
To reconstruct a 3D shape with the proposed representation of an input observation $o\in\mathcal{X}$ (e.g., single image),
we design a DNN architecture that receives the input and outputs the corresponding matrix $H$, center $c$, and parameter $R$ for each primitive as shown in Figure~\ref{Fig:net}.

\subsubsection{Training losses}
We apply various losses to reconstruct 3D shapes.

\noindent
\textbf{Loss sign.}
Since the target surface in 3D space divides the inside and outside, 
we define a sign function $\sign(x,y,z):\mathbb{R}^3\longrightarrow\{0,-1,1\}$ on sample points $P\subset\mathbb{R}^3$ where the values $0$, $-1$, and $1$ correspond to the points on the target surface, its inside, and its outside, respectively.
Likewise, we can classify the points on/inside/outside the reconstructed implicit surfaces $\mathcal{S}(x,y,z)$ and reduce a loss between their predicted and ground truth signs.
We use mean square error (MSE) as the loss function as Eq.~\ref{Eq:loss_points}:
\begin{equation}\label{Eq:loss_points}
\begin{split}
    \mathcal{L}_{\mathcal{B}}^{sign}=\sum_{\mathclap{i\in\{on,in,out\}}}{\lambda_i\cdot\mathbb{E}_{\textbf{p}_i\sim P}\lVert\tanh({\mathcal{S}(\textbf{p}_i))-\sign(\textbf{p}_i)} \rVert^2},
\end{split}
\end{equation}
where $\mathcal{B}\subset\mathcal{X}$ and $\lambda_{i}$ are a training batch and the weights corresponding to each sign, respectively.
Note that $\mathcal{L}_{\mathcal{B}}^{sign}$ enforce the network to reconstruct the desired surface and refuse to generate the redundant surfaces simultaneously.
Moreover, the MSE loss forces further attention on distinguishing the inside and outside points near the reconstructed surface because their $\tanh{\mathcal{S}(\textbf{p}_i)}$ are near zero while their ground truths are $-1$ and $+1$, respectively.

\noindent 
\textbf{Loss normal.}
To improve the reconstruction, we use the normal vectors as second-order information. Therefore, we define a MSE loss between the ground-truth normal vectors of the sample points $\textbf{p}_{on}$ on the surface of a target mesh model $\textbf{n}_{\text{g}}$ and their normal vectors $\textbf{n}_{\text{r}}$ obtained by Eq.~\ref{Eq:loss_normal}:
\begin{equation}\label{Eq:loss_normal}
\begin{split}
    \mathcal{L}^{n}_{\mathcal{B}}=\mathbb{E}_{\textbf{p}_{on}\sim P}\lVert  \textbf{n}_{r}(\textbf{p}_{on}) - \textbf{n}_{g}(\textbf{p}_{on})\rVert^2,
\end{split}
\end{equation}
where the normal vectors on the union surface can be directly determined for any point on the surface as Eq.~\ref{Eq:normal_vector}: 
\begin{equation}\label{Eq:normal_vector}
\begin{split}
    \textbf{n}_r=&\frac{\nabla \mathcal{S}(x,y,z)}{||\nabla \mathcal{S}(x,y,z)||}=\frac{(\frac{\partial \mathcal{S}}{\partial x},\frac{\partial \mathcal{S}}{\partial x},\frac{\partial \mathcal{S}}{\partial z})}{\sqrt{(\frac{\partial \mathcal{S}}{\partial x})^2+(\frac{\partial \mathcal{S}}{\partial y})^2+(\frac{\partial \mathcal{S}}{\partial z})^2}},\\
    &\text{subject to:}
    \quad \nabla \mathcal{S}(x,y,z) = \nabla p_{m^{\ast}}(x,y,z), \\
    &\hspace{15mm} \quad m^{\ast}= \argmin_m(p_m(x,y,z)).
\end{split}
\end{equation}
Note that $m^{\ast}$ is the index of the closest primitive (i.e., the primitive with the smallest value $p_{m^{\ast}}(x,y,z)$) to the point.

Finally, the total loss $\mathcal{L}^{\text{total}}_\mathcal{B}$ is the weighting average of all defined losses with the corresponding weights as Eq.~\ref{Eq:loss_total}:
\begin{equation}\label{Eq:loss_total}
    \mathcal{L}^{\text{total}}_\mathcal{B} = \mathcal{L}^{\sign}_{\mathcal{B}}+\lambda_n\mathcal{L}^{n}_{\mathcal{B}}.
\end{equation}
\subsubsection{Implementation details}\label{section:implementation_details}
We consider the bounding box $\mathcal{C}=[-1,1]^3$ and fit the given input 3D shapes into it by keeping its aspect ratio. Then we extract $1M$ points from $[-1.1,1.1]^3\in\mathbb{R}^3$ surrounding the 3D shapes, and at each iteration, we randomly select $1\%$ of them as jointly inside points $\textbf{p}_{in}$ and outside points $\textbf{p}_{out}$ for all shapes in the batch $\mathcal{B}$.
Moreover, we pick $10k$ points $\textbf{p}_{on}$ on the surface of each 3D shape and randomly select $20\%$ of them at each iteration. 

To capture the information of the input observation $o\in\mathcal{X}$ we employ the pretrained ResNet-18~\cite{ResNet} as the encoder. Then three sets of independent fully connected (FC) layers $(4096,4096,4096,55\times M),(1024,512,256,3\times M),(256,M)$ decodes the encoded features to obtain the parameters of each symmetric matrix $B$, scalar $R$ and center $c$ for $M=100$ primitives in Eq.~\ref{Eq:union}, respectively. All FC layers except the last layers are empowered by the $ReLU$ non-linear activation function. We also apply three batch normalization layers after the first three FC layers of $B$ decoder to accelerate training and boost the performance.

To ensure $H$ is a PD matrix, we parameterize it as  Eq.~\ref{Eq:pd}:
\begin{equation}\label{Eq:pd}
\begin{split}
    H=BB^T+\alpha I\succ 0
\end{split}
\end{equation}
where $\alpha=0.0001$ is a small scalar factor, and $B$ and $I$ are $10\times 10$ symmetric and identity matrices, respectively. 
Moreover, we apply the $sigmoid$ function on the output of the $R$ decoder to generate a value in $(0,1)\subset\mathbb{R}^3$ as the parameter $R$ of Eq.~\ref{Eq:bound_primitive} to control the size of all primitives and guarantee that they are not larger than the size of the bounding box $\mathcal{C}$.
Furthermore, we apply $\tanh$ on the output of the $c$ decoder to generate the centers inside the bounding box $\mathcal{C}$.
Therefore, each primitive is parameterized with 59 parameters in total including three parameter for the center $c=(c_1,c_2,c_3)$, one parameter for the $R$, and 55 parameters for the matrix $B$.  
 
We set the parameters $\lambda_{on}$, $\lambda_{in}$, $\lambda_{out}$, and $\lambda_{n}$ as 2, 1, 10, and 1, respectively.  
We train our encoder-decoder architecture with Adam optimizer with the initial learning rate 1e-4, weight decay 1e-7, and batch size of 64.
We implement our model in Python3.7 using PyTorch via CUDA instruction.

\begin{figure*}[t]
\begin{center}
        \begin{subfigure}[b]{0.64\textwidth}
               \includegraphics[width=\textwidth]{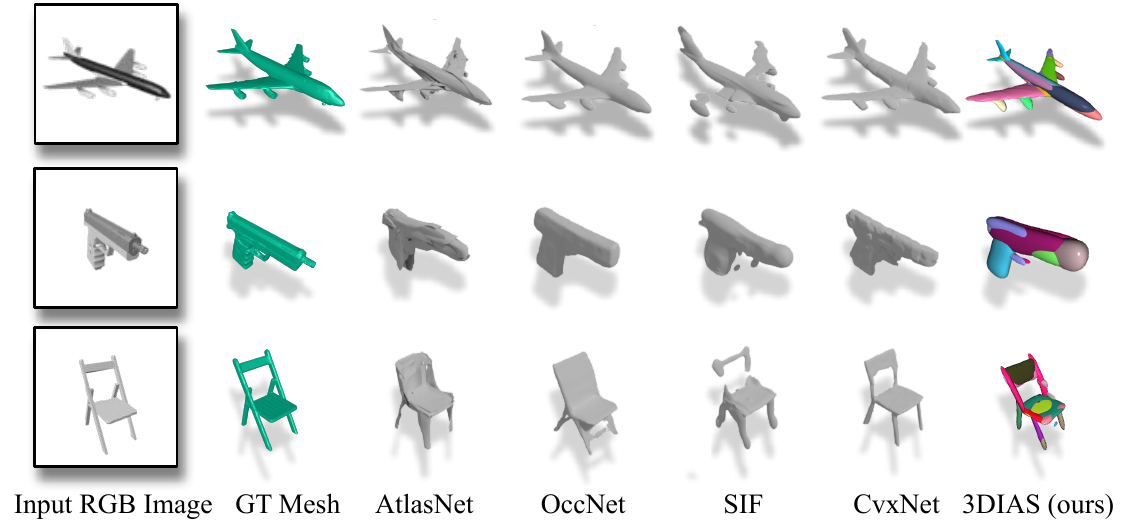}
               \caption{}
                \label{fig:qualitative_all}
        \end{subfigure}%
        \qquad
        \begin{subfigure}[b]{0.31\textwidth}
                \includegraphics[width=\linewidth]{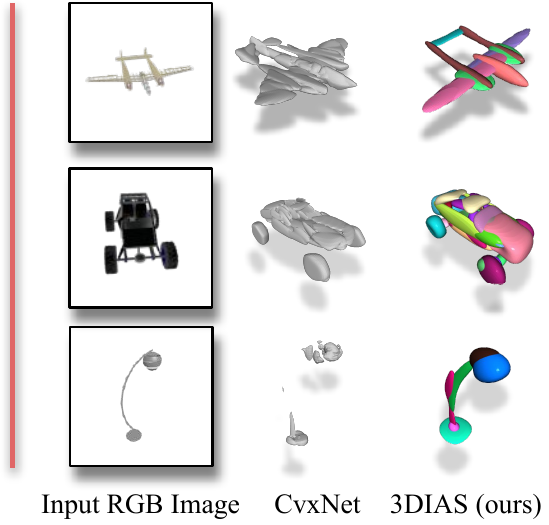}
                \caption{}
                \label{fig:qualitative_cvxnet}
        \end{subfigure}
\end{center}
   \caption{Qualitative comparison on single RGB image 3D shape reconstruction. SIF~\cite{Genova2019LearningST}, AtlasNet~\cite{DBLP:journals/corr/abs-1802-05384}, OccNet\cite{ONet}, CvxNet\cite{deng2020cvxnet}, and our 3DIAS output reconstructed 3D shape from the given RGB image. (a) Comparison with other methods for the samples shown in CvxNet~\cite{deng2020cvxnet}. (b) More qualitative comparisons with CvxNet~\cite{deng2020cvxnet}.} 
   \label{Fig:qualitative}
\end{figure*}

\section{Experiments}

\begin{table*}
\begin{center}
\renewcommand{\arraystretch}{1.0}
\resizebox{\textwidth}{!}{\begin{tabular}{|l|cccccc|cccccc|ccccc|}
\hline
\multirow{2}{*}{Category} 
&\multicolumn{6}{c|}{IoU}
&\multicolumn{6}{c|}{Chamfer}
&\multicolumn{5}{c|}{F-Score}\\
\cline{2-18}
& P2M & AtlasNet & OccNet & SIF & CvxNet & 3DIAS & P2M & AtlasNet & OccNet & SIF & CvxNet & 3DIAS & AtlasNet & OccNet & SIF & CvxNet & 3DIAS
\\ \hline\hline
airplane & {0.420}  & {-}  & {0.571}  & {0.530}  & \textbf{0.598}  & {0.549}  & {0.187}  & {0.104}  & {0.147}  & {0.167}  & {0.093} & \textbf{0.087} & {67.24}  & {62.87}  & {52.81}  & \textbf{68.16} & {59.48}\\
bench & {0.323}  & {-}  & \textbf{0.485}   & {0.333}  & {0.461}  & \textbf{0.485}  & {0.201}  & {0.138}  & {0.155}  & {0.261}  & {0.133}  & \textbf{0.106}  & {54.50}  & {56.91}  & {37.31}  & {54.64}  & \textbf{60.17} \\
cabinet & {0.664}  & {-}  & \textbf{0.733}  & {0.648}  & {0.709}  & {0.730}  & {0.196}  & {0.175}  & {0.167}  & {0.233}  & {0.160}  & \textbf{0.123}  & {46.43}  & {61.79}  & {31.68}  & {46.09}  & \textbf{61.81} \\
car & {0.552}  & {-}  & \textbf{0.737}  & {0.657}  & {0.675}  & \textbf{0.737}  & {0.180}  & {0.141}  & {0.159}  & {0.161}  & {0.103}  & \textbf{0.091}  & {51.51}  & {56.91}  & {37.66}  & {47.33}  & \textbf{58.07} \\
chair & {0.396}  & {-}  & {0.501}  & {0.389}  & {0.491}  & \textbf{0.509}  & {0.265}  & {0.209}  & {0.228}  & {0.380}  & {0.337}  & \textbf{0.186}  & {38.89}  & {42.41}  & {26.90}  & {38.49}  & \textbf{43.14}\\
display & {0.490}  & {-}  & {0.471}  & {0.491}  & \textbf{0.576}  & {0.538}  & {0.239}  & \textbf{0.198}  & {0.278}  & {0.401}  & {0.223}  & {0.211}  & \textbf{42.79}  & {38.96}  & {27.22}  & {40.69}  & {42.40} \\
lamp & {0.323}  & {-}  & {0.371}  & {0.260}  & {0.311}  & \textbf{0.381}  & {0.308}  & \textbf{0.305}  & {0.479}  & {1.096}  & {0.795}  & {0.607}  & {33.04}  & \textbf{38.35}  & {20.59}  & {31.41}  & {37.52} \\
speaker & {0.599}  & {-}  & \textbf{0.647}  & {0.577}  & {0.620}  & {0.638}  & {0.285}  & \textbf{0.245}  & {0.300}  & {0.554}  & {0.462}  & {0.351}  & {35.75}  & \textbf{42.48}  & {22.42}  & {29.45}  & {39.16}\\
rifle & {0.402}  & {-}  & {0.474}  & {0.463}  & \textbf{0.515}  & {0.423}  & {0.164}  & {0.115}  & {0.141}  & {0.193}  & \textbf{0.106}  & {0.116}  & \textbf{64.22}  & {56.52}  & {53.20}  & {63.74}  & {47.44} \\
sofa & {0.613}  & {-}  & {0.680}  & {0.606}  & {0.677}  & \textbf{0.685}  & {0.212}  & {0.177}  & {0.194}  & {0.272}  & {0.164}  & \textbf{0.158}  & {43.46}  & {48.62}  & {30.94}  & {42.11}  & \textbf{49.73} \\
table & {0.395}  & {-}  & {0.506}  & {0.372}  & {0.473}  & \textbf{0.509}  & {0.218}  & {0.190}  & \textbf{0.189}  & {0.454}  & {0.358}  & {0.245}  & {44.93}  & \textbf{58.49}  & {30.78}  & {48.10}  & {57.63} \\
phone & {0.661}  & {-}  & {0.720}  & {0.658}  & {0.719}  & \textbf{0.751}  & {0.149}  & {0.128}  & {0.140}  & {0.159}  & {0.083}  & \textbf{0.080}  & {58.85}  & {66.09}  & {45.61}  & {59.64}  & \textbf{71.35} \\
vessel & {0.397}  & {-}  & {0.530}  & {0.502}  & \textbf{0.552}  & {0.538}  & {0.212}  & \textbf{0.151}  & {0.218}  & {0.208}  & {0.173}  & {0.206}  & \textbf{49.87}  & {42.37}  & {36.04}  & {45.88}  & {40.70} \\
\hline
mean & {0.480}  & {-}  & {0.571}  & {0.499}  & {0.567}  & \textbf{0.575}  & {0.216}  & \textbf{0.175}  & {0.215}  & {0.349}  & {0.245}  & {0.197}  & {48.57}  & {51.75}  & {34.86}  & {47.36}  & \textbf{52.22}  \\
\hline
\end{tabular}}
\end{center}
\caption{Evaluation of single image 3D shape reconstruction. We evaluate and compare our method (3DIAS) to the state-of-the-art methods including P2M~\cite{DBLP:journals/corr/abs-1804-01654}, AtlasNet~\cite{DBLP:journals/corr/abs-1802-05384}, OccNet\cite{ONet}, SIF~\cite{Genova2019LearningST}, and CvxNet\cite{deng2020cvxnet} on a part of ShapeNet dataset~\cite{DBLP:journals/corr/ChangFGHHLSSSSX15} in terms of IoU, Chamfer, and F-score.}
\label{table:RGB}
\end{table*}

In this section, we provide information about the evaluation setups and show qualitative and quantitative results of our method compared to state-of-the-art methods on single RGB image 3D shape reconstruction.
We also perform various ablation studies to analyze our method better.
More experiments are available in the supplementary material.

\subsection{Dataset and Metrics}
We evaluate our approach on the subset of the ShapeNet dataset ~\cite{DBLP:journals/corr/ChangFGHHLSSSSX15} with the same image renderings and training/testing split provided by Choy \etal~\cite{DBLP:journals/corr/ChoyXGCS16}. 
We also employ mesh-fusion~\cite{Stutz2018ARXIV} to generate watertight meshes from the 3D CAD models. We then use Houdini~\cite{houdini} to extract the inside/outside/on points and the normal vectors.
For evaluation, we use the volumetric IoU, Chamfer ~\cite{DBLP:journals/corr/FanSG16}, and F-Score ~\cite{10.1145/3072959.3073599} metrics.
Volumetric IoU is used to measure the overlapped volume between the ground truth meshes and the reconstructed surfaces.
Chamfer is the mean of the accuracy and the completeness score. The mean distance of points on the reconstructed surface to their nearest neighbors on the ground truth mesh is defined as the accuracy metric. The completeness metric is defined in the opposite direction of the accuracy metric.  F-score is the harmonic mean of precision which shows the percentage of correctly reconstructed surfaces. To compute IoU, we sample $100k$ points from the bounding box. To evaluate Chamfer and F-score, we first transfer the reconstructed surfaces to meshes, then similar to CvxNet\cite{deng2020cvxnet} we sample $100k$ points on the reconstructed and the ground-truth meshes.

\subsection{Reconstruction}
We experimentally evaluate our method 3DIAS trained on multi-class and compare it with state-of-the-art methods on single RGB image 3D shape reconstruction and summarize the results in Table~\ref{table:RGB}. 
The experiments demonstrate the superiority of 3DIAS compared to the explicit-based methods P2M~\cite{DBLP:journals/corr/abs-1804-01654} and AtlasNet~\cite{DBLP:journals/corr/abs-1802-05384}, the isosurface-based method OccNet~\cite{ONet}, and the recent primitive-based methods SIF~\cite{Genova2019LearningST} and CvxNet~\cite{deng2020cvxnet} in terms of volumetric IoU and F-score. We also achieve the second-best performance with the Chamfer metric. We show more quantitative results of 3DIAS for the trained network on single-class in the supplementary material.

Moreover, we qualitatively evaluate 3DIAS trained on single-class and compare it with the previous methods in Figure~\ref{Fig:qualitative}. The results illustrate that 3DIAS achieves smooth surfaces with desirable geometrical details. 3DIAS, unlike the previous methods, is successful in reconstructing the 3D shape with more complex topologies (e.g., chair) as shown in Figure~\ref{fig:qualitative_all}. Moreover, compared to CvxNet~\cite{deng2020cvxnet}, 3DIAS can better reconstruct thin shapes (e.g., lamp) and when similar shapes are rare in the training dataset (e.g., airplane and car), see Figure~\ref{fig:qualitative_cvxnet}.

\subsection{Ablation study}
We perform several ablation studies to analyze our proposed representation and reconstruction procedure. First, we show the ability of our method to generate more complex primitives compared to other primitive-based methods. Then we compare the required number of parameters to represent 3D shapes with our representation and other methods. Finally, we demonstrate the power of our reconstruction scheme to learn the semantic structures in an unsupervised manner. Moreover, we evaluate the effects of the designed constraints and the defined loss functions in reconstructing 3D shapes with high detailed appearances.

\begin{figure}[!h]
\begin{center}
\includegraphics[width=\linewidth]{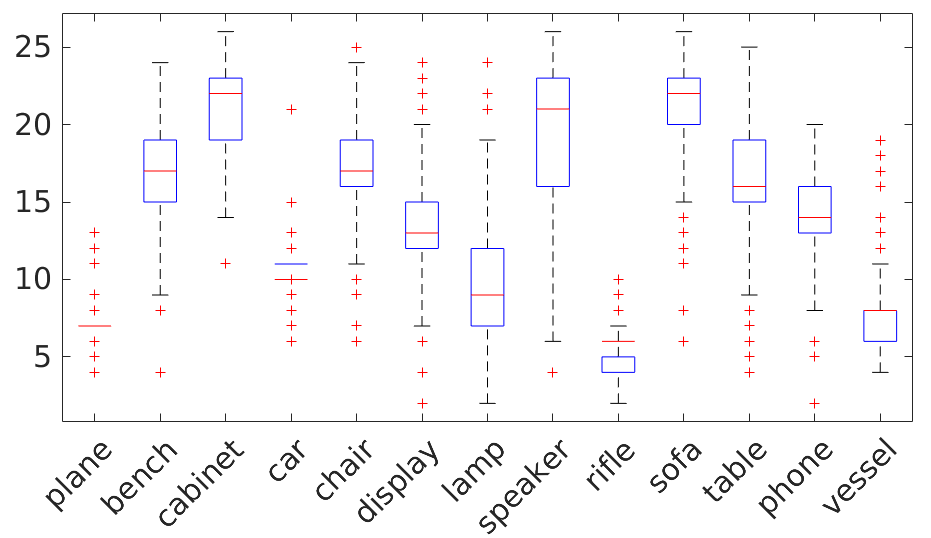}
\end{center}
   \caption{Statistics of the number of primitives. We compute the average number of primitives selected among all $M=100$ primitives by the network for each category.}
   \label{Fig:efficiency}
\end{figure}
\begin{table} 
\begin{center}
\begin{tabular}{|l|c|c|c|c|}
\hline
Representation & SIF & OccNet & CvxNet & 3DIAS\\
\hline\hline
Num of params & 700 & 11M & 7700 &  480\\
\hline
\end{tabular}
\end{center}
\caption{Number of parameters. The average number of parameters for representing 3D shapes by different methods.} 
\label{table:numOFparams}
\end{table}

\subsubsection{Complexity of primitives}
\begin{figure}[!t]
        \begin{subfigure}[b]{0.27\textwidth}
                \includegraphics[width=\linewidth, page=1]{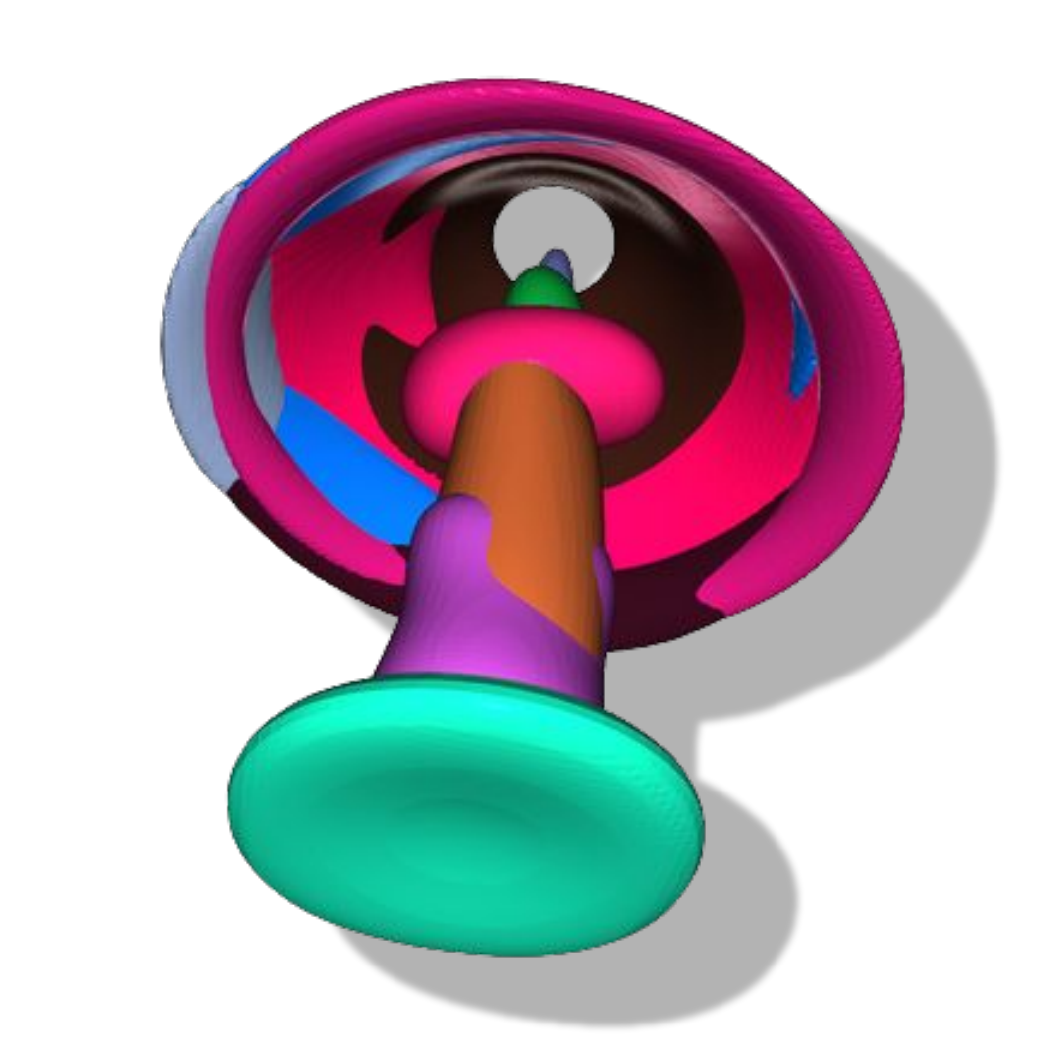}
                \includegraphics[width=\linewidth, page=2]{primitive_complex.pdf}
                \caption{Lamp}
                \label{fig:complex_lamp}
        \end{subfigure}%
        \qquad
        \begin{subfigure}[b]{0.27\textwidth}
                \includegraphics[width=\linewidth, page=3]{primitive_complex.pdf}
                \includegraphics[width=\linewidth, page=4]{primitive_complex.pdf}
                \caption{Speaker}
                \label{fig:complex_speaker}
        \end{subfigure}
        \qquad
        \begin{subfigure}[b]{0.27\textwidth}
                \includegraphics[width=\linewidth, page=5]{primitive_complex.pdf}
                \includegraphics[width=\linewidth, page=6]{primitive_complex.pdf}
                \caption{Chair}
                \label{fig:complex_chair}
        \end{subfigure}
        \caption{The complexity of our primitives. The first and the second rows show the reconstructed shapes and their corresponding primitives, respectively. The proposed primitive can effectively present curved and torus shapes.}\label{fig:complex_prim}
\end{figure}

We illustrate that our proposed constrained primitive is able to form more geometrical (e.g., curved) and topological (e.g., genus-one) complex shapes as shown in Figure~\ref{fig:complex_prim}. While the previous primitive-based methods such as cubes~\cite{Cubes}, ellipsoids~\cite{Genova2019LearningST}, superquadrics~\cite{Superquadrics}, and convexes~\cite{deng2020cvxnet} cannot form such complex shapes.

\subsubsection{The number of parameters}
In section~\ref{section:3D_shape_reprsentation} we argue that a primitive may have no real solution when $|p^{3\downarrow}(x,y,z)|<|p^{4}(x,y,z)|$ or $|p^{3\downarrow}(x,y,z)|$ is non-negative for all points $(x,y,z)\in\mathbb{R}^3$ (i.e., no valid surface). Accordingly, our method can ignore some of the primitives among all $M=100$ primitives by assigning positive definite coefficient matrices $A$ to them and maintain a sufficient number of primitives. 
Therefore, these not-solvable primitives do not participate in reconstructing the surface $\mathcal{S}$. 
Note that, our method selects sets of primitives that are mainly different for inter-category shapes and have a large overlap for intra-category shapes. 
During the test phase, we can efficiently check the eigenvalues for the coefficient matrix of each primitive and eliminate the primitives with non-negative eigenvalues.
Our experiments demonstrate that our network selects few primitives to reconstruct 3D shapes, as shown in Figure~\ref{Fig:efficiency}.
In addition, since each quartic primitive can finally be identified with only $35$ coefficients $a_{ijk}$, the surface $\mathcal{S}$ of 3D shapes with 3DIAS representation can be represented with only $35\times 13.71\simeq480$ number of parameters on average.
3DIAS requires $68.571\%$, $0.004\%$, and $6.234\%$ of the parameters used in SIF~\cite{Genova2019LearningST}, OccNet~\cite{ONet}, and CvxNet~\cite{deng2020cvxnet} on average to represent 3D shapes, respectively, see Table~\ref{table:numOFparams}.



\subsubsection{Unsupervised semantics segmentation}
\begin{figure}[t]
\begin{center}
\begin{minipage}[t]{.45\linewidth}
\vspace{0pt}
\centering
    \includegraphics[width=\linewidth]{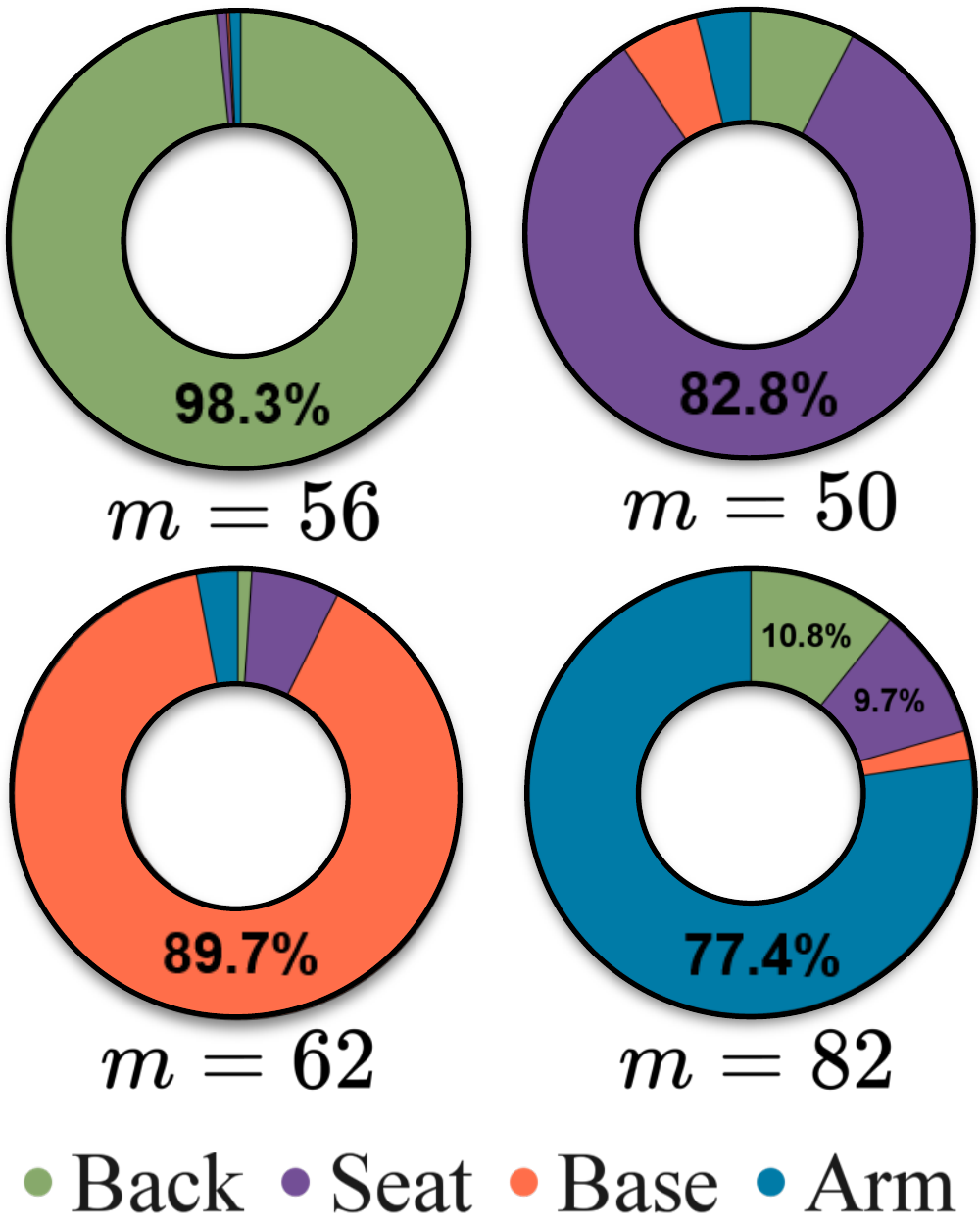}
\end{minipage}%
\qquad
\begin{minipage}[t]{.45\linewidth}
\vspace{20pt}
\centering
\setlength\tabcolsep{0.7pt}
\begin{tabular}[b]{|l|c|c|c|}
\hline
\small
\multirow{2}{*}{Part} &
\multicolumn{3}{c|}{Accuracy}\\
\cline{2-4}
& BAE
& CvxNet
& 3DIAS\\
\hline\hline 
\textcolor{asparagus}{Back} & {86.36} & {\bf 91.50} & {88.87} \\
\textcolor{darklavender}{Seat} & {73.66} & {\bf 90.63} & {70.29} \\
\textcolor{outrageousorange}{Base} & {\bf 88.46} & {71.95} & {78.51} \\
\textcolor{cerulean}{Arm} & {65.75} & {38.94} & {\bf 74.86} \\
\hline
mean  & {\bf 78.56} & {73.25} & {78.13}  \\
\hline
\end{tabular}
   \caption{Evaluation of semantic segmentation. (left) The distribution of PartNet labels within 4 primitives in chair class. (right) The classification accuracy for each part. We follow the evaluation method introduced in cvxnet~\cite{deng2020cvxnet}.}
   \label{Fig:seman_statis}
   \end{minipage}
\end{center}
\end{figure}

\begin{figure}[t]
    \centering
    \begin{subfigure}[b]{0.45\linewidth}        
        \centering
        \includegraphics[width=\linewidth, page=1]{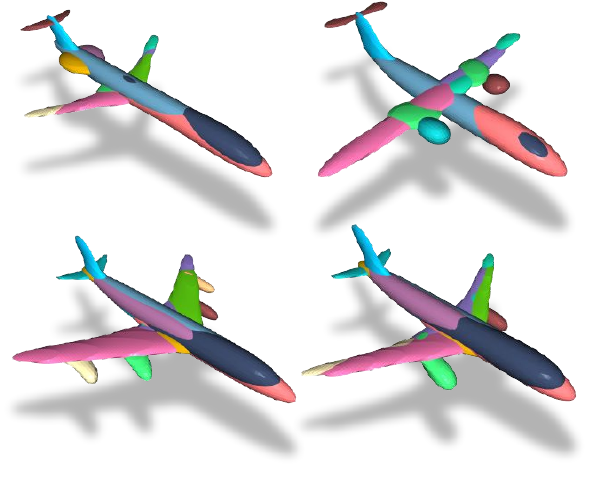}
        \caption{Airplane}
        \label{fig:sem_airplane}
    \end{subfigure}
    \qquad
    \begin{subfigure}[b]{0.45\linewidth}        
        \centering
        \includegraphics[width=\linewidth, page=2]{sem_total_2row.pdf}
        \caption{Chair}
        \label{fig:sem_chair}
    \end{subfigure}
    \caption{Qualitative results on unsupervised semantic segmentation. We visualize the results of 3DIAS for some samples in the categories of (a) airplane and (b) chair.}
    \label{Fig:Qualitative_semantics}
\end{figure}

We also illustrate that our network learns a semantic structure without any part-level supervision such that one primitive usually covers the same part of reconstructed 3D shapes in the same class with 3DIAS representation.
We evaluate the semantic structures on the PartNet\cite{PartNet} dataset having the labels of hierarchical parts of the shapeNet.
The quantitative experiments in Figure~\ref{Fig:seman_statis} show that our method achieves better and comparable average accuracy compared to CvxNet~\cite{deng2020cvxnet} and BAE~\cite{BAE}, respectively.  In addition, 3DIAS achieves better accuracy than both methods for thin parts (e.g., arm).
Moreover, our qualitative experiments illustrate that one primitive tends to cover the same semantic part as shown in Figure~\ref{Fig:Qualitative_semantics}. This tendency is more pronounced for the dominant primitive that covers more points. For instance, the dominant primitives mainly cover the seat of chairs because most of the chairs have seat parts. Please see the supplementary material for more examples.

\subsubsection{Effects of constraints}
We study the effect of our designed constraints on reconstructing 3D shapes.
In each experiment, we evaluate or baseline by ignoring one or more constraints.
The quantitative results based on volumetric IoU, Chamfer, and F-Score show the importance of each constraint, see Table~\ref{table:constraints}. 
We believe these constraints encourage the network to reconstruct a detailed 3D shape, especially the center constraint.

\begin{table}[t]
\small
\begin{center}
\setlength\tabcolsep{4.5pt} 
\begin{tabular}{|l|c|c|c|}
\hline
Constraints
& IoU
& Chamfer
& F-Score\\
\hline\hline
-center & {0.549} & {0.387} & {46.72} \\
-scale & {0.559} & {0.261} & {48.45} \\
-scale, -closedness & {0.546} & {0.280} & {44.44} \\
\hline
All  & {\bf 0.575} & {\bf 0.197} & {\bf 52.22}  \\
\hline
\end{tabular}
\end{center}
\caption{Ablation study on constraints. We compare the effects of the center, the scale, and the closedness constraints in terms of IoU, Chamfer, and F-score. Note that in each configuration we ignore one or two constraints.}
\label{table:constraints}
\end{table}

\subsubsection{Effects of losses}
While $\mathcal{L}^{sign}$ for the inside/outside points tries to distinguish inside and outside of 3D shapes, it is not enough to achieve a detailed surface due to the lack of sample points near the surface. Therefore, points on the surface and their normal vectors can facilitate the reconstruction. Note that normal vectors carry important information on 3D geometry, such as the local orientation of surfaces. 
Accordingly, we use $\mathcal{L}^{sign}$ loss and $\mathcal{L}^{n}$ loss for the points on the surface to better approximate the surfaces.
We evaluate the effects of each $\mathcal{L}^{sign}$, $\mathcal{L}^{n}$, and their combination by excluding them for the points on the surface and summarize the results in Table~\ref{table:losses}.   
Note that for all the experiments in Table~\ref{table:losses} we do not exclude $\mathcal{L}^{sign}$ for the inside/outside points.
The results indicate the importance of points on the surface to achieve more detailed 3D shapes.

\begin{table}[t]
\small
\begin{center}
\setlength\tabcolsep{4.5pt} 
\begin{tabular}{|l|c|c|c|}
\hline
Losses
& IoU
& Chamfer
& F-Score\\
\hline\hline
-$\mathcal{L}^{n}$ & {0.568} & {0.210} & {49.17} \\
-$\mathcal{L}^{sign}$ & {0.548} & {0.219} & {43.77} \\
-$\mathcal{L}^{sign}$, -$\mathcal{L}^{n}$ & {0.542} & {0.232} & {42.37} \\
\hline
All  & {\bf 0.575} & {\bf 0.197} & {\bf 52.22} \\
\hline
\end{tabular}
\end{center}
\caption{Ablation study on losses. We compare the effects of $\mathcal{L}^{sign}$ and $\mathcal{L}^{n}$ losses for the points on the surface in terms of IoU, Chamfer, and F-score. Note that in each configuration we ignore one or two loss functions. Moreover, we do not exclude the $\mathcal{L}^{sign}$ for the inside/outside points. Please see the supplementary material for more examples.}
\label{table:losses}
\end{table}

\section{Conclusion}
In this paper, we propose a primitive-based representation and a learning scheme in which the primitives are learnable implicit algebraic surfaces that can jointly approximate 3D shapes.
We design various constraints and loss functions to achieve high-quality and detailed 3D shapes. 
We experimentally demonstrate that our method outperforms state-of-the-art methods in most of the metrics.
Moreover, we illustrate that our method can learn semantic meanings without part-level supervision by automatically selecting sets of primitives parametrized by only a few parameters.
In the future, we will utilize the solvability of the designed primitives to develop a soft renderer which leads to reconstruct the 3D shapes with self-supervised learning.

\section*{Acknowledgement}
This work was supported in part by an IITP grant funded by the Korean government [No. 2021-0-01343, Artificial Intelligence Graduate School Program (Seoul National University)].

{\small
\bibliographystyle{ieee_fullname}
\bibliography{ms}
}

\iccvfinalcopy 

\def\iccvPaperID{5644} 
\def\httilde{\mbox{\tt\raisebox{-.5ex}{\symbol{126}}}}

\ificcvfinal\pagestyle{empty}\fi

\title{Supplementary Material \textit{for} \\ \vspace{2mm} 3DIAS: 3D Shape Reconstruction with Implicit Algebraic Surfaces}

\author{Mohsen Yavartanoo$^{\ast}$ \qquad Jaeyoung Chung$^{\ast}$ \qquad Reyhaneh Neshatavar \qquad Kyoung Mu Lee \\ASRI, Department of ECE, Seoul National University, Seoul, Korea\\
{\tt\small \{myavartanoo,robot0321,reyhanehneshat,kyoungmu\}@snu.ac.kr}}

\maketitle
\ificcvfinal\thispagestyle{empty}\fi

\renewcommand{\thetable}{S1}
\begin{table}[t]
\small
\begin{center}
\setlength\tabcolsep{2pt}
\begin{tabular}{|l|cc|cc|cc|cc|}
\hline
\multirow{2}{*}{Category} 
&\multicolumn{2}{c|}{IoU}
&\multicolumn{2}{c|}{Chamfer}
&\multicolumn{2}{c|}{F-Score}
&\multicolumn{2}{c|}{\#primitive}\\
\cline{2-9}
& Multi & Single & Multi  & Single & Multi & Single & Multi & Single
\\ \hline\hline
airplane & {0.549} & {\bf 0.621} & {\bf 0.087} & {0.580} & {59.48} & {\bf 69.51} & {6.915} & {\bf 23.42} \\
bench & {\bf 0.485} & {0.462} & {\bf 0.106} & {0.677} & {\bf 60.17} & {59.60}  & {16.97} & {\bf 22.24} \\
cabinet & {\bf 0.730} & {0.726} & {\bf 0.123} & {0.141} & {\bf 61.81} & {53.24} & {20.91} & {\bf 22.98} \\
car & {0.737} & {\bf 0.747} & {0.091} & {\bf 0.082} & {58.07} & {\bf 61.08}  & {10.02} & {\bf 37.96} \\
chair & {\bf 0.509} & {0.493} & {\bf 0.186} & {0.304} & {\bf 43.14} & {38.85}  & {17.17} & {\bf 19.15} \\
display & {\bf 0.538} & {0.511} & {\bf 0.211} & {1.137} & {\bf 42.40} & {37.85} & {13.88} & {\bf 14.95} \\
lamp & {\bf 0.381} & {0.352} & {\bf 0.607} & {1.494} & {\bf 37.52} & {35.96}  & {9.946} & {\bf 17.54} \\
speaker & {\bf 0.638} & {0.632} & {0.351} & {\bf 0.310} & {\bf 39.16} & {34.19} & {19.24} & {\bf 19.65} \\
rifle & {0.423} & {\bf 0.509} & {\bf 0.116} & {0.760} & {47.44} & {\bf 61.60} & {4.719} & {\bf 17.16} \\
sofa & {\bf 0.685} & {0.667} & {\bf 0.158} & {0.186} & {\bf 49.73} & {45.20}  & {21.35} & {\bf 24.55} \\
table & {\bf 0.509} & {0.478} & {\bf 0.245} & {0.369} & {\bf 57.63} & {50.96}  & {15.97} & {\bf 16.68} \\
phone & {\bf 0.751} & {0.734} & {\bf 0.080} & {0.168} & {\bf 71.35} & {65.85} & {14.11} & {\bf 18.23} \\
vessel & {0.538} & {\bf 0.550} & {0.206} & {\bf 0.200} & {40.70} & {\bf 43.89}  & {7.067} & {\bf 19.41} \\
\hline
mean &  {\bf 0.575} & {\bf 0.575} & {\bf 0.197} & {0.493} & {\bf 52.22} & {50.60} & {13.71} & {\bf 21.07} \\
\hline
\end{tabular}
\end{center}
\caption{Comparison of multi-class and single-class single RGB image 3D shape reconstruction in terms of IoU, Chamfer, F-score, and the number of primitives.}
\label{table:single-multi}
\end{table}

\begin{figure*}
\centering
    \begin{subfigure}{\linewidth}
        \includegraphics[page=1 ,width=\linewidth]{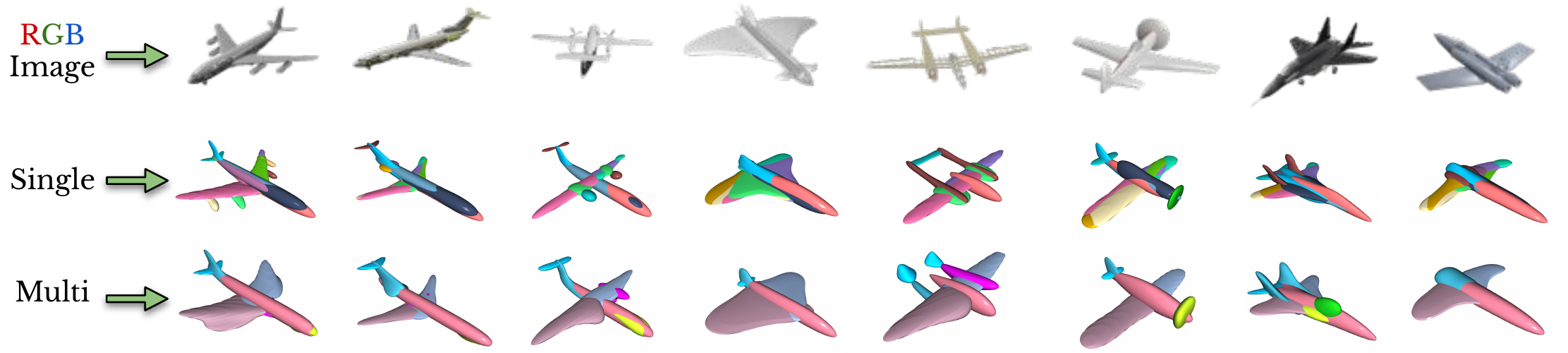}
            \caption*{Airplane}
    \end{subfigure}


    \begin{subfigure}{\linewidth}
        \includegraphics[page=2 ,width=\linewidth]{suppsinglemulti.pdf}
            \caption*{Bench}
    \end{subfigure}


    \begin{subfigure}{\linewidth}
        \includegraphics[page=3 ,width=\linewidth]{suppsinglemulti.pdf}
            \caption*{Cabinet}
    \end{subfigure}


    \begin{subfigure}{\linewidth}
        \includegraphics[page=4 ,width=\linewidth]{suppsinglemulti.pdf}
            \caption*{Car}
    \end{subfigure}

    \begin{subfigure}{\linewidth}
        \includegraphics[page=5 ,width=\linewidth]{suppsinglemulti.pdf}
            \caption*{Chair}
    \end{subfigure}
\end{figure*}
\begin{figure*}
\centering

    \begin{subfigure}{\linewidth}
        \includegraphics[page=6 ,width=\linewidth]{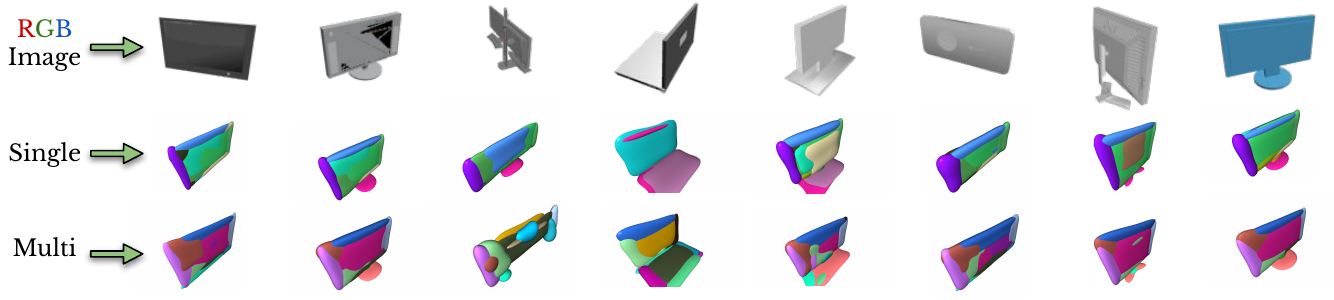}
            \caption*{Display}
    \end{subfigure}

    \begin{subfigure}{\linewidth}
        \includegraphics[page=7 ,width=\linewidth]{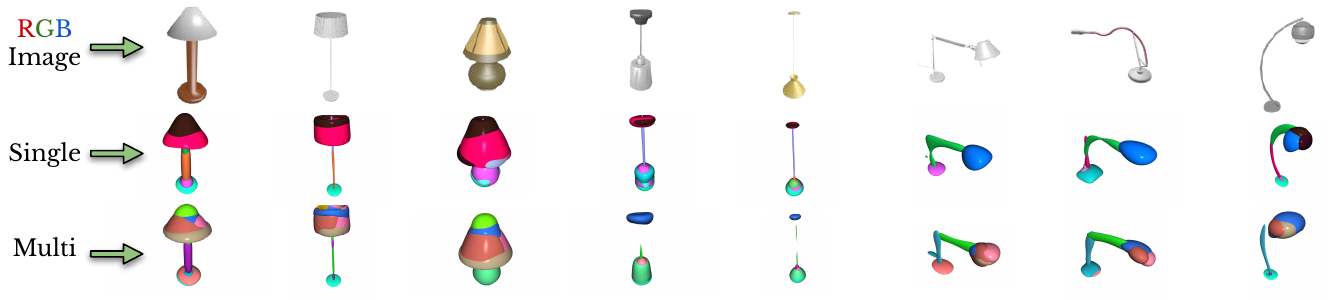}
            \caption*{Lamp}
    \end{subfigure}

    \begin{subfigure}{\linewidth}
        \includegraphics[page=8 ,width=\linewidth]{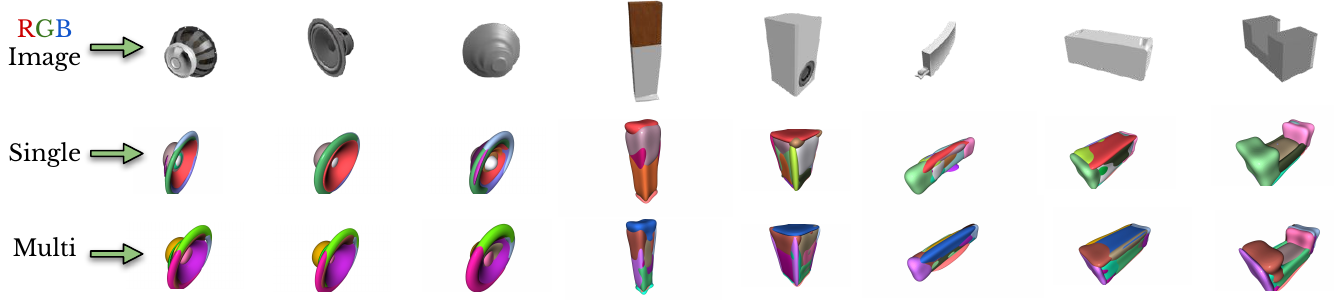}
            \caption*{Speaker}
    \end{subfigure}

    \begin{subfigure}{\linewidth}
        \includegraphics[page=9 ,width=\linewidth]{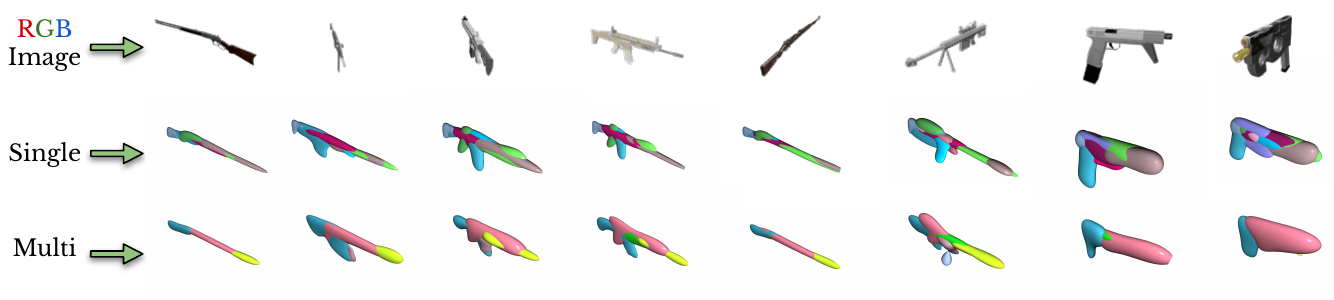}
            \caption*{Rifle}
    \end{subfigure}

    \begin{subfigure}{\linewidth}
        \includegraphics[page=10 ,width=\linewidth]{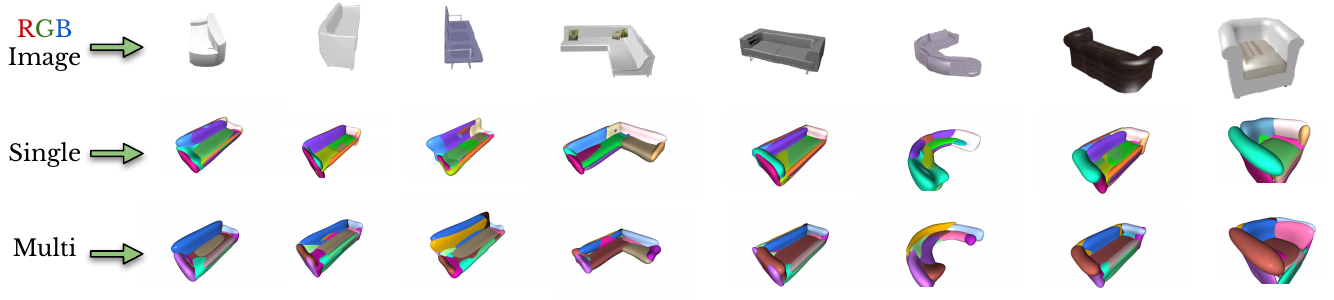}
            \caption*{Sofa}
    \end{subfigure}
\end{figure*}
\renewcommand{\thefigure}{S2}
\begin{figure*}
\centering
    \begin{subfigure}{\linewidth}
        \includegraphics[page=11 ,width=\linewidth]{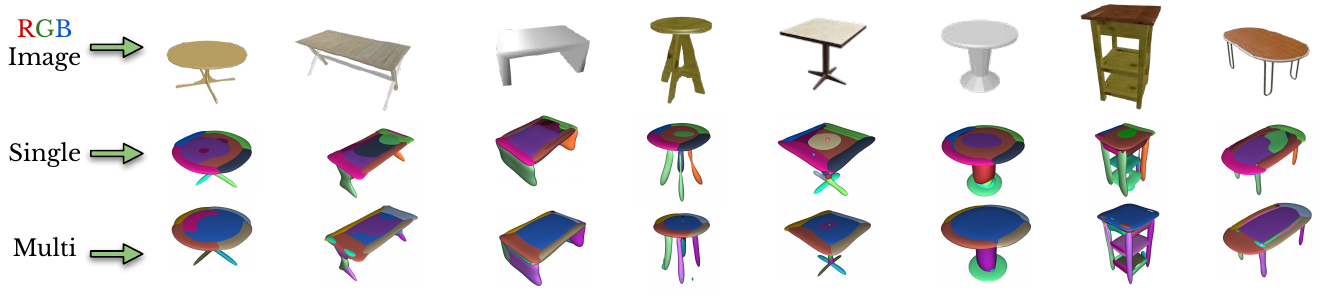}
            \caption*{Table}
    \end{subfigure}

    \begin{subfigure}{\linewidth}
        \includegraphics[page=12 ,width=\linewidth]{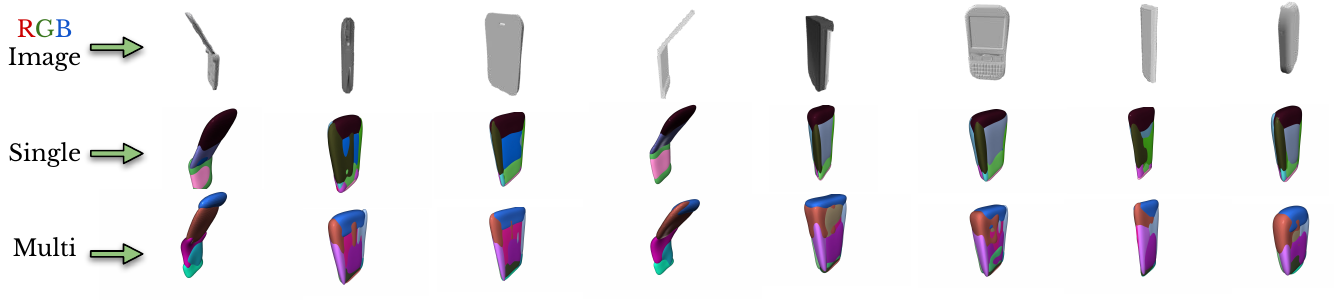}
            \caption*{Phone}
    \end{subfigure}

    \begin{subfigure}{\linewidth}
        \includegraphics[page=13 ,width=\linewidth]{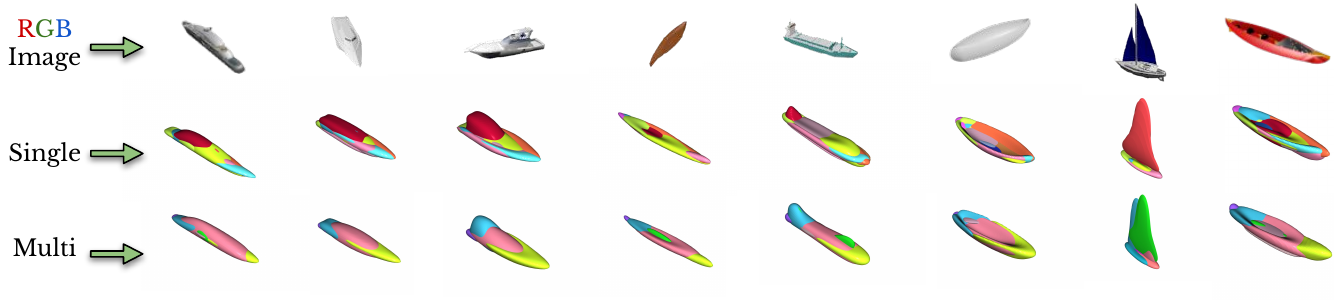}
            \caption*{Vessel}
    \end{subfigure}
\caption{Qualitative comparison of single-class and multi-class cases. We visualize the results for some samples in each category. The first, second, and third lines for each category show the input RGB images, single-class training results, and multi-class training results, respectively.}
    \label{fig:suppsinglemulti}
\end{figure*}

\appendixpageoff
\appendixtitleoff

\begin{appendices}

  \setcounter{section}{18}

\section{3DIAS analysis}
In this section, we provide more experimental analyses of our proposed 3DIAS method.
First, we prove that our scale constraint satisfies the criteria of the closedness constraint. Then we quantitatively and qualitatively show and compare our experiments of 3DIAS trained on single-class and multi-class.
Finally, we show the effect of viewpoint in reconstructing 3D shapes with our proposed method.

\subsection{Holding closedness constraint}
In the section~\textcolor{red}{3.1.2}, we claim that the parametrized coefficient matrix $A$ holds the criteria for the closedness constraint. To ensure the closedness constraint is satisfied, the coefficient matrix $A_{[5:10]}$ in Eq.~\textcolor{red}{3} of the fourth-degree terms $p^4(x,y,z)$ must be positive definite.
The matrix $A_{[5:10]}$, as the principal sub-matrix of the coefficient matrix $A$ of $p(x,y,z)$, is the summation of the corresponding sub-matrices $H_{[5:10]}$ and $Q_{[5:10]}$. 
Since the matrix $H$ is assumed positive definite, its principal sub-matrices (e.g., $H_{[5:10]}$) are also positive definite. Moreover, the corresponding sub-matrix $Q_{[5:10]}$ of the coefficient matrix $Q$ is a diagonal matrix with the values $[1,1,1,0,0,0]$; hence it is positive semi-definite. Accordingly, the sub-matrix $A_{[5:10]}$ as the summation of a positive definite matrix and a positive semi-definite matrix is also positive definite, which satisfies the closedness constraint.

\subsection{Multi-class vs single-class training}
We also evaluate our method for the trained network individually on each class and compare the results in terms of IoU, Chamfer, and F-score with the multi-class case and summarize the results in Table~\ref{table:single-multi}.
Surprisingly, the comparison demonstrates that our method trained on the multi-class can better reconstruct on average, presumably due to the overfitting to the training set.
However, our method trained on the single-class selects more primitives on average to reconstruct 3D shapes as shown in Table~\ref{table:single-multi}. We believe it is because more available primitives are per class to represent 3D shapes in the single-class training, while the network must distribute the limited primitives among all categories in the multi-class training. 

We also visualize the reconstructed 3D shapes by 3DIAS for both the single-class and the multi-class cases in Figure~\ref{fig:suppsinglemulti}. 
The results show that the primitives share the same semantically meaning among the 3D shapes in the same category in both single-class and multi-class cases. 
Based on our qualitative and quantitative experiments, we believe that the Chamfer is not a suitable and reliable metric for the 3D shape reconstruction task. For instance, Chamfer shows better performance for multi-class training on airplane and rifle categories, while the qualitative results show better appearances for single-class training.

\subsection{Effect of viewpoint}
We also show the effect of viewpoint in Figure~\textcolor{red}{S3}.
The results illustrate that the quality of the reconstructed 3D shapes is highly dependent on the point of view when similar shapes are rare in the training dataset (e.g., bottom airplane). However, we show that this effect decreases when there are enough similar shapes to the query shape (e.g., top airplane) in the training dataset.

\renewcommand{\thefigure}{S3}
\begin{figure*}
\begin{center}
    \centering
    \includegraphics[width=\linewidth]{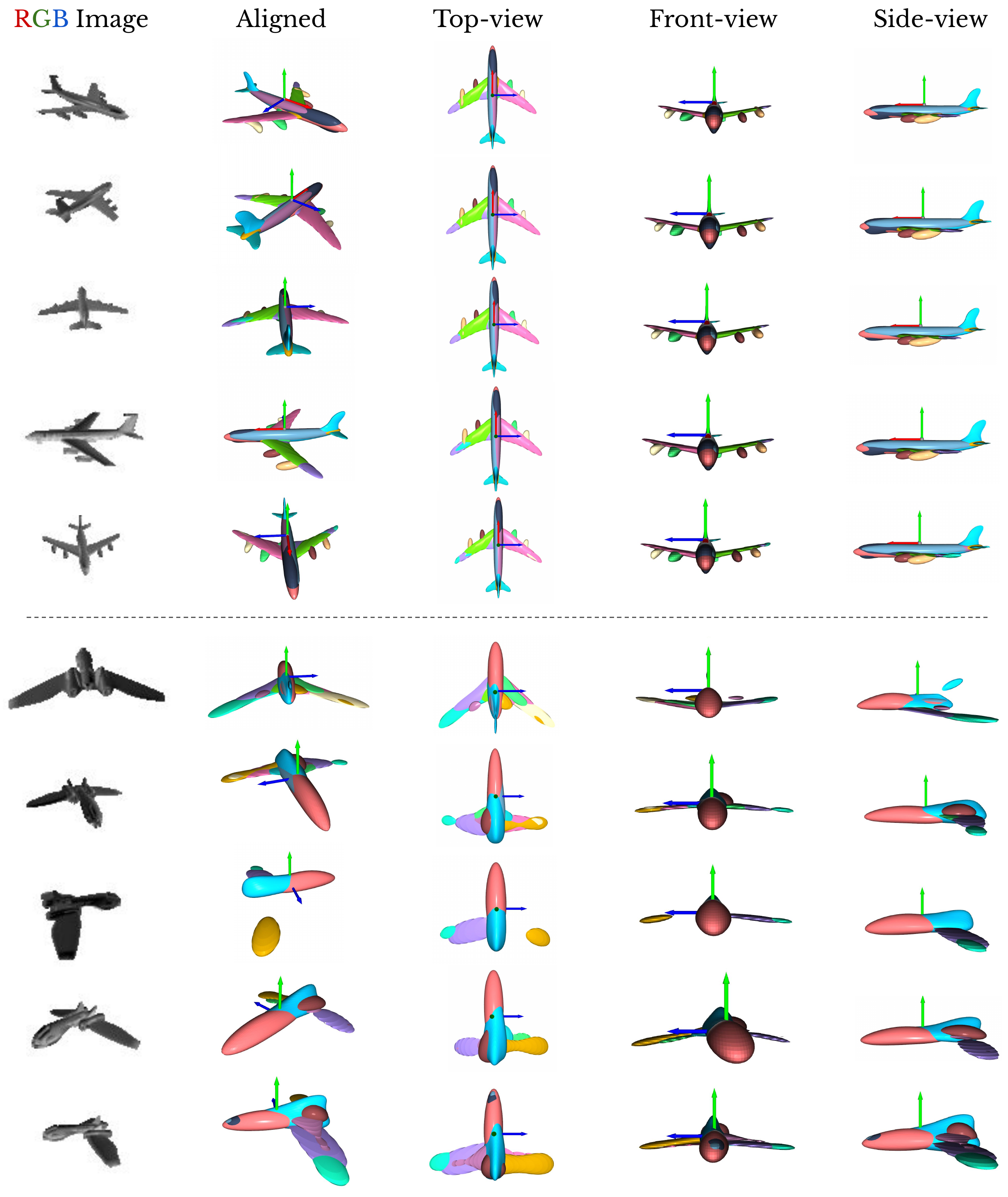}
    \captionof{figure}{Effect of viewpoint in 3D shape reconstruction. We visualize the reconstructed 3D shape by 3DIAS for two samples. (top) and (bottom) show two airplanes with small and large viewpoint effects on their appearances, respectively.}
\end{center}%
\label{fig:viewpoints}
\end{figure*}

\end{appendices}

\end{document}